\documentclass[10pt,twocolumn,letterpaper]{article}

\usepackage[final,pagenumbers]{cvpr}  % Camera-ready version with page numbers

\newcommand{\red}[1]{{\color{red}#1}}

\renewcommand{\red}[1]{#1}

\definecolor{iccvblue}{rgb}{0.21,0.49,0.74}
\usepackage[pagebackref,breaklinks,colorlinks,allcolors=iccvblue]{hyperref}
\usepackage{caption}
\usepackage{cuted}

\usepackage{multirow}
\usepackage{tabularx}
\usepackage{graphicx} % for \resizebox

\usepackage[most]{tcolorbox}
\definecolor{takeaway}{rgb}{0.97,0.94,0.91}
\newtcbox{\takeawaybox}{on line,
  colback=takeaway,
  colframe=takeaway,
  boxrule=0pt,
  arc=2pt,       % corner roundness
  boxsep=1pt,    % padding
  left=3pt,right=3pt,top=1pt,bottom=1pt
}
\newtcolorbox{takeawayblock}{
  colback=takeaway,
  colframe=takeaway,
  boxrule=0pt,
  arc=2pt,
  left=2pt,right=2pt,top=2pt,bottom=2pt,
  boxsep=0pt,
  before skip=1pt,
  after skip=1pt
}
\def\confName{CVPR}
\def\confYear{2026}

\title{Ablate-to-Validate: Are Vision-Language Models Really Using Continuous Thought Tokens?}

\author{Tianyi Zhang$^*$ \quad Mahtab Bigverdi$^*$ \quad Ranjay Krishna\\
University of Washington\\
{\tt\small \{tzhang26, mahtab, ranjay\}@cs.washington.edu}\\
{\small $^*$Equal contribution.}
}

\begin{document}
\maketitle

% \begin{strip}
% \vspace{-2.0em}
% \centering
% \includegraphics[width=\textwidth]{figures/ATV_figure.pdf}
% \captionof{figure}{
% Overview of the Token Replacement Test (TRT). Given a fixed image and prompt, TRT replaces the intermediate thought-token span using zero, random, first-repeat, or oracle/ground-truth tokens while keeping the prompt, image, token budget, and decoding procedure fixed. Comparing original, random, and oracle performance isolates whether auxiliary-token gains come from genuine content utilization or from incidental span-position and token-budget effects.
% }
% \label{fig:trt_method}
% \end{strip}

\begin{strip}
\vspace{-3.0em}
\centering
\begin{minipage}{\textwidth}
\centering
\includegraphics[
  width=0.88\linewidth,
  trim=6pt 6pt 6pt 6pt,
  clip
]{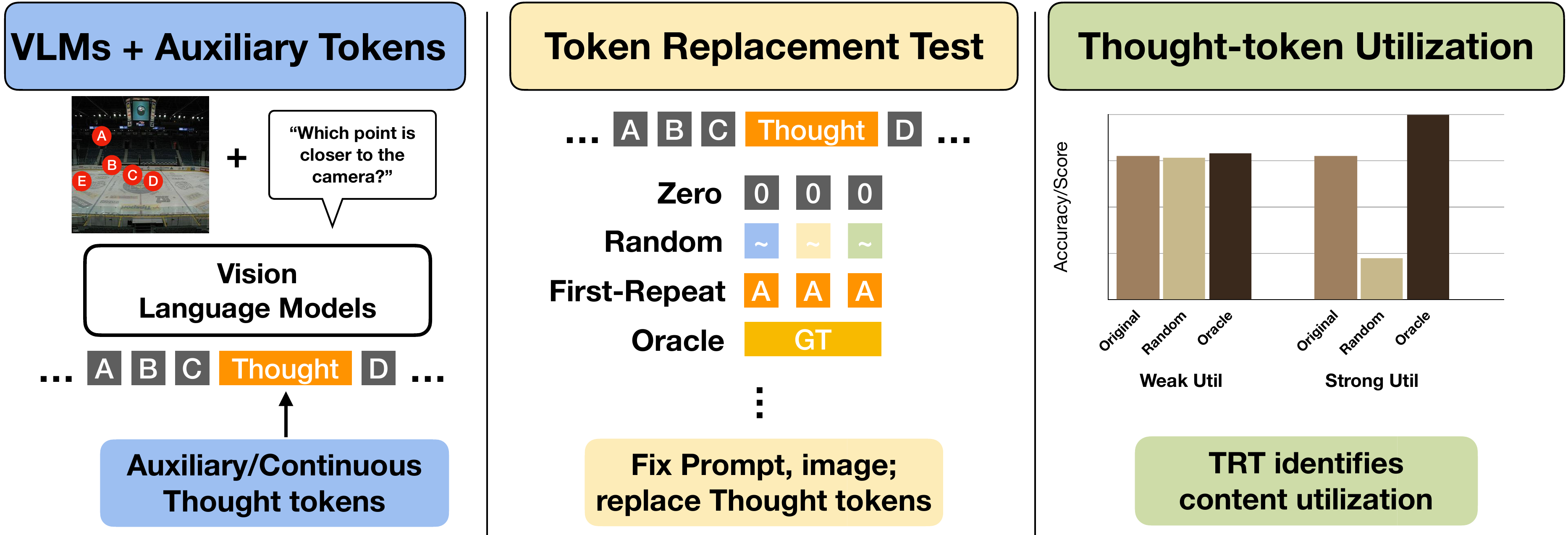}

\vspace{0.2em}
\captionof{figure}{
\small
Overview of the Token Replacement Test (TRT). TRT replaces the intermediate thought-token span while fixing the prompt, image, token budget, and decoding procedure. Comparing identity, random, first-repeat, and oracle replacements separates content utilization from span-position and token-budget effects.
}
\label{fig:trt_method}
\end{minipage}
\vspace{-1.0em}
\end{strip}

\begin{abstract}
   Vision language models (VLMs) are increasingly augmented with continuous or latent non-textual tokens intended to support ``visual thinking.'' Despite the improved task accuracy, this alone does not show that models actually use these tokens for reasoning. 
Gains may instead arise from confounds such as added context length, special-token anchoring, or training-time regularization. 
We formalize a diagnostic principle, Ablate-to-Validate, for testing whether latent-token content is genuinely utilized. 
We instantiate this principle as the Token Replacement Test (TRT), a standardized suite of content-replacement ablations. 
TRT measures whether performance depends on the information carried by latent tokens rather than their mere presence. It probes (1) \textit{span-position bias} through zero and random replacement, (2) disentangles \textit{token-budget from token-diversity} effects through first-repeat and count-matched variants, and (3) evaluates information content using oracle or \textit{ground-truth token injection} together with distribution-matched random baselines. 
As a controlled testbed, we study relative depth reasoning, where continuous depth embeddings can be inserted at explicit span positions under a fixed token budget. 
We train and prove LLaVA and Qwen2.5-VL models both to predict and to consume these tokens, and show that TRT applies across heterogeneous latent-token model backbones.
\red{Concretely, we cover two trained backbones (LLaVA-13B, Qwen2.5-VL-3B) with continuous and discrete depth spans across three frozen visual encoders (SigLIP2, CLIP, DINOv2) and multiple token budgets, and additionally apply TRT to three off-the-shelf visual-thinking systems (Mirage, Mull-Tokens, CoVT) evaluated on BLINK, VSP, and CV-Bench.}
Our results show that accuracy gains can be a misleading proxy for latent-token reasoning: across multiple model backbones, types of continuous visual tokens, and compute budgets, VLMs retain most of the improvement even when latent-token content is corrupted or replaced, revealing a persistent gap between ``having a latent channel'' and actually using it as an information bottleneck.
By separating true content utilization from alternative explanations, TRT provides a simple and standardized way to evaluate continuous thought tokens in vision-language models, and we recommend reporting such diagnostics as standard practice. \textit{Code will be released at the \href{https://tjazhang.github.io/ablate_to_validate}{project page}.}

\end{abstract}
\section{introduction}
\label{sec:introduction}

% \begin{figure}[t]
%     \centering
%     \includegraphics[width=\textwidth]{figures/ATV_figure.pdf}
%     \caption{  An overview of the Token Replacement Test (TRT) designed to evaluate thought-token utilization in Vision Language Models (VLMs). As shown, the model processes an image containing specific points (A, B, C, D, E) alongside a user query such as ``Which point is closer to the camera?''. The VLM outputs a sequence that includes continuous visual tokens. In the TRT methodology, the prompt and image are kept fixed, but the intermediate thought tokens are systematically replaced. The test utilizes several replacement strategies, including Zero, Random, First-Repeat, and Oracle/Ground Truth (GT). By comparing the final accuracy/score of the Original output  against the Random and Oracle replacements, TRT reveals whether the model exhibits Weak Util or Strong Util of the token information. Ultimately, TRT successfully identifies true content utilization.}
%     \label{fig:trt_method}
% \end{figure}

Vision-language models (VLMs) first rose to prominence on tasks where success largely reduces to conditioning on visual features and producing short-form text: captioning, visual question answering, and referring expression comprehension~\cite{vinyals2015showandtell,antol2015vqa,kazemzadeh2014refcoco}. In these regimes, the dominant bottlenecks are perceptual coverage and alignment. In other words, is the model seeing the right thing and attaching the right words. For such tasks, scaling data, architectures, and instruction tuning tends to yield steady improvements.

But computer vision did not begin as ``language-conditioned pattern recognition.'' Early vision work treated images as projections of underlying structure, and sought intermediate representations that make geometry and objecthood explicit~\cite{do1961machine,minsky1969introduction,marr2010vision}. That original framing is resurfacing in modern evaluations. Recent benchmarks increasingly target settings where perception is necessary but not sufficient, and \textit{multi-step visual reasoning} becomes the limiting factor: expert-level multimodal understanding~\cite{yue2023mmmu}, mathematical reasoning grounded in images~\cite{lu2023mathvista}, and integrated capability tests that combine perception with knowledge, spatial reasoning, and computation~\cite{yu2023mmvet}. In parallel, the language modeling community has shown that explicitly eliciting intermediate structure, most notably via \textit{chain-of-thought}~\cite{wei2022cot}, can substantially improve performance on hard problems~\cite{wei2022cot,kojima2022zeroshotcot}. Together, these trends raise a basic question for multimodal systems: what should intermediate structure look like when plain text is an awkward—or lossy—interface for visual computation?

A natural answer is to augment VLMs with the ability to reason beyond language, using intermediate visual representations. Ideally, they would can emit and later consume such intermediate representations as part of their reasoning process. One family of such representations uses \textit{discrete} visual tokens; for instance, ``perception tokens'' encode intermediate artifacts like depth maps or bounding boxes, and conditions reasoning on those tokens~\cite{bigverdi2025perception}. A second family uses continuous or latent non-textual tokens, often framed as \textit{continuous} visual thoughts, that the model produces and later reads~\cite{tong2024metamorph,ray2025mulltokens,qin2025covt,yang2025mirage}. 
Across both families, these visual token mechanisms routinely report accuracy gains. 
However, accuracy gains can reflect incidental effects rather than information in the tokens: a fixed insertion position can become a reliable cue, a larger effective context/capacity from extra tokens can provide more computational strength, or auxiliary objectives for learning visual tokens might regularize training.

This paper asks a more pointed version of the question: \textit{ when auxiliary tokens help, are gains driven by the information in the tokens, or by incidental position and budget effects?} We propose \textbf{Ablate-to-Validate}, an empirical principle for answering this. If a model truly relies on auxiliary-token information, then systematically perturbing token values, while holding the prompt, token budget, image features, and decoding procedure fixed, should induce predictable performance changes. 
We operationalize this principle with the \textbf{Token Replacement Test (TRT)}, our standardized inference-time replacement protocol that isolates (1) \textit{span-position} bias via zero and distribution-matched random replacement, (2) \textit{token-budget versus token-diversity} effects via first-repeat and count-matched variants when token counts differ, and (3) genuine information utilization via oracle or \textit{ground-truth injection} compared against matched random baselines (Figure~\ref{fig:trt_method}).

To make these interventions precise, we develop a controlled testbed based on questions that require depth reasoning. Depth provides a continuous, geometrically grounded signal that is both meaningful for visual computation and amenable to clean counterfactual manipulations: we can zero or randomize embeddings, and we can define a clear upper bound by injecting ground-truth (oracle) depth embeddings. Concretely, we insert an explicit depth-token span into the language stream and train models to both predict and consume continuous depth embeddings at those positions. We implement this pipeline for multiple VLM backbones (LLaVA~\cite{liu2023llava} and Qwen2.5-VL~\cite{bai2025qwen25vl}), enforcing a fixed visual token budget across experiments. We also compare against a discrete depth baseline~\cite{bigverdi2025perception} that quantizes depth into a VQVAE codebook and trains via cross-entropy under the same token budget. Finally, to show TRT’s broader applicability, we apply the same ablate-to-validate lens to additional off-the-shelf VLMs: Mirage \cite{yang2025mirage}, Mull-Tokens \cite{ray2025mulltokens}, and Chain-of-Visual-Thought (CoVT) \cite{qin2025covt} with visual token mechanisms.

Empirically, TRT reveals a recurring content--utility gap: across many configurations, continuous visual tokens function more like a fixed interface than an information-rich reasoning channel. Across multiple model backbones, types of continuous visual tokens, and compute budgets, replacing the model's predicted tokens with zeros or distribution-matched random vectors produces little to no degradation, and injecting oracle tokens yields only marginal gains in some cases. This pattern indicates that models can profit from the presence and placement of a visual-token span---its budget, its anchoring, and its training-time effects---without reliably decoding the span's semantic content as a bottleneck for the final decision. TRT provides a minimal, standardized validation step to separate these cases from settings where auxiliary token content is genuinely used, and we recommend reporting TRT-style diagnostics as standard practice when introducing new latent-token mechanisms.

\section{Related work}
\label{sec:related-work}
We situate our work amongst the growing body of VLM algorithms that inject the capability of models to reason beyond language.

\noindent\textbf{How VLMs process visual content.}
Early VLMs such as Flamingo~\cite{alayrac2022flamingo} showed that inserting cross-modal attention into a (largely) frozen LLM can yield strong few-shot transfer, while later open models such as LLaVA~\cite{liu2023llava} and Qwen2.5-VL~\cite{bai2025qwen25vl} popularized simpler projection-based designs that map visual features into the LLM embedding space. These architectures achieve strong performance on many perception-heavy benchmarks, and scaling data and model sizes have often yielded steady improvements.

As evaluation has shifted toward multi-step reasoning regimes, research has explored whether explicitly eliciting intermediate structure improves multimodal performance. Multiple efforts have tried to include intermediate rationales or structured traces to support visual--linguistic inference~\cite{zhang2023multimodal,lu2022learn}. However, serializing geometric or perceptual computation into text can be awkward and lossy: viewpoint changes, occlusions, and metric comparisons are often more naturally expressed in a visual or continuous representation than in natural language. This motivates approaches that augment VLMs with visual token streams intended to serve as intermediate reasoning substrates.

\noindent\textbf{Spatial and depth reasoning benchmarks as testbeds.}
A large body of work diagnoses where VLMs fail on spatial reasoning, revealing brittleness even for seemingly basic predicates. SpatialSense~\cite{yang2019spatialsense}, VSR~\cite{liu2022vsr}, and What'sUp~\cite{kamath2023whatsup} use adversarial or minimal-pair designs to reduce language priors and expose persistent failures in left/right and above/below reasoning. Recent benchmarks extend this to 3D and viewpoint sensitivity: 3DSRBench~\cite{ma20253dsrbench} shows that models break under modest viewpoint changes, while ViewSpatial-Bench~\cite{li2025viewspatialbench} highlights a perspective gap between camera-centered and human-centered viewpoints. For multi-image/video settings, VSI-Bench~\cite{yang2025thinkinginspace} and MMSI-Bench~\cite{yang2025mmsibench} test whether models can maintain a consistent spatial map over time and viewpoints, and MindCube~\cite{yin2025mindcube} targets spatial mental modeling from limited views. Related benchmark methodology work further emphasizes that shortcuts remain pervasive and motivates debiasing through iterative filtering~\cite{brown2025trainontestset}.

These benchmarks are invaluable for locating failure modes, but they typically do not isolate \emph{how} a model internally computes spatial conclusions. In particular, they do not provide standardized diagnostics for whether a proposed intermediate mechanism (e.g., visual token streams) is genuinely \emph{used as information} versus merely acting as a training scaffold. Our work complements these evaluations by introducing a controlled, intervention-based diagnostic (TRT) and instantiating it in a depth reasoning testbed where counterfactual token replacements and oracle injections are well-defined. Our method can be expanded to also support spatial benchmarks.
% \noindent\textbf{Visual externalization and multimodal traces.}
% A complementary direction externalizes intermediate computation using external tools that produce explicit visual traces. Visual Sketchpad~\cite{hu2024visualsketchpad} equips models with drawing actions for iterative refinement, while MVoT~\cite{li2025mvot} argues that interleaving visualization with text reasoning can help on dynamic spatial tasks. ThinkMorph~\cite{gu2025thinkmorph} similarly studies interleaved text--image reasoning traces. These approaches emphasize interpretability and controllability of intermediate steps, but they do not directly answer whether \emph{latent} visual channels are treated by models as information-bearing bottlenecks when present.
% \paragraph{Visual externalization and multimodal traces.}
% A complementary direction externalizes intermediate computation using tools or generated visual traces, such as sketching, visualization-of-thought, or interleaved text--image reasoning~\cite{hu2024visualsketchpad,li2025mvot,gu2025thinkmorph}. These methods improve interpretability and control, but do not directly test whether latent visual channels are consumed as information-bearing bottlenecks.

\noindent\textbf{Discrete visual tokens for reasoning.}
One line of work introduces explicit intermediate artifacts as tokens and conditions subsequent reasoning on them. Perception tokens~\cite{bigverdi2025perception} exemplify this approach by encoding intermediate representations such as depth maps or bounding boxes into discrete tokens that the model can generate and then consume during inference. More broadly, discrete tokenizations provide a natural interface to autoregressive decoding and allow straightforward accounting of token budgets.

\noindent\textbf{Latent and continuous visual tokens.}
Motivated by the hypothesis that continuous representations better preserve perceptual structure than discretized alternatives, several recent methods augment VLMs with latent or continuous token streams that lie outside the standard vocabulary. MetaMorph~\cite{tong2024metamorph} introduces interleaved continuous visual tokens in the generation stream and trains models for multimodal understanding and generation. Chain-of-Visual-Thought (CoVT)~\cite{qin2025covt} distills information from multiple vision experts into a compact continuous span that the VLM predicts before generating answers. Mirage~\cite{yang2025mirage} frames latent tokens as ``machine mental imagery,'' recasting hidden states into a latent visual channel through a two-stage training paradigm. Mull-Tokens~\cite{ray2025mulltokens} generalizes this idea to modality-agnostic latent thinking tokens, and Latent Implicit Visual Reasoning~\cite{li2025latent} trains models to discover visual reasoning tokens without explicit supervision.

Across these works, visual token mechanisms often report improvements on a variety of perception and reasoning benchmarks. However, accuracy gains alone do not establish that a model relies on \emph{visual token content}. Improvements can arise from confounds such as fixed span placement (a reliable marker), changes in effective capacity via token budget, or visual objectives that regularize training. Our work directly targets this gap: we propose the Token Replacement Test (TRT), a standardized content-replacement diagnostic that separates (i) span-position bias, (ii) token-budget vs.\ token-diversity effects, and (iii) genuine information utilization via oracle injection baselines.

\section{Method}

\subsection{Preliminaries: visual tokens in VLMs}

Autoregressive VLMs generate an output sequence conditioned on an image $I$ and a textual prompt $x$.
We write the textual output as $y_{1:T}$ with $y_t \in \mathcal{V}$, and the next-token distribution
\begin{equation}
p(y_t \mid y_{<t}, x, I)=\mathrm{Softmax}(W h_t),
\end{equation}
where $h_t \in \mathbb{R}^H$ is the hidden state and $W$ is the output projection.

A growing class of methods augments this process with an \emph{visual token span} intended to act as an intermediate representation. We denote the visual span by $u_{1:K}$ where $K$ is the visual token budget. Depending on the method, the visual tokens may be
\begin{equation}
u_i \in 
\begin{cases}
\mathcal{V}_u & \text{(discrete visual tokens)}\\
% V_u is visual voabulary
\mathbb{R}^D & \text{(continuous/latent visual tokens)}.
\end{cases}
\end{equation}
Operationally, the model first produces (or is given) a visual span at designated positions, and then generates downstream text conditioned on that span. This creates a natural distinction between \textbf{prediction} and \textbf{utilization}: a model may learn to \emph{predict} visual tokens (e.g., via a visual loss) without \emph{using their content} when producing the final answer. Since most prior work reports only end-task accuracy, it remains unclear whether visual tokens are acting as an information-bearing bottleneck or merely as a scaffold (e.g., a fixed marker, extra budget, or regularization).

Our goal is to test \emph{utilization}: does changing the \emph{values} of $u_{1:K}$ (while keeping everything else fixed) change downstream performance in the way we would expect if the model were relying on visual content?

\subsection{Ablate-to-Validate Principle}

We propose \textbf{Ablate-to-Validate}: if a model genuinely relies on the information encoded in a visual span, then systematically perturbing that span at inference should induce predictable changes in the final output quality. We instantiate this principle as the \textbf{Token Replacement Test (TRT)}, a standardized suite of \emph{content-replacement interventions} applied to the visual span $u_{1:K}$ while holding fixed the prompt $x$, image features from $I$, the token budget $K$, and the decoding procedure.

\subsection{Token Replacement Test (TRT)}
We treat the visual span as an explicit object in the forward pass. Let $\mathrm{Inject}(x, u_{1:K})$ denote the augmented input sequence (or internal stream) in which the visual span is consumed. TRT evaluates a model by running the same example under a collection of replacements $u_{1:K}\leftarrow \tilde{u}_{1:K}$ and measuring the change in task performance.

\paragraph{Replacement interventions.}
TRT uses the following replacements, designed to isolate distinct confounds:
\begin{itemize}
\item \textbf{Identity re-injection.} $\tilde{u}_{1:K} \leftarrow u_{1:K}$, verifying that the interception and replacement pathway is correct.
\item \textbf{Zero replacement.} $\tilde{u}_{1:K} \leftarrow 0$ (or a designated null token for discrete spans), testing sensitivity to removing visual content.
\item \textbf{Random replacement.} $\tilde{u}_{1:K} \leftarrow \epsilon$, with $\epsilon$ sampled i.i.d.\ (e.g., Gaussian for continuous spans), testing whether performance depends on the specific predicted content.
\item \textbf{Distribution-matched random.} $\tilde{u}_{1:K} \leftarrow \epsilon$ sampled to match the empirical distribution of the model's predicted visual tokens (or ground-truth tokens when available), controlling for scale and marginal statistics.
\item \textbf{First-repeat.} $\tilde{u}_i \leftarrow u_1$ for all $i$, preserving budget and span placement while removing token diversity.
\item \textbf{Count-matched variants.} When comparing methods whose native visual spans differ in length, we apply the same replacement while matching the number of intervened tokens to the number used by the corresponding identity run for that example.\footnote{Some methods do not enforce a fixed auxiliary-token budget during generation. For example, in Mirage \cite{yang2025mirage} the number of auxiliary tokens can vary across examples and, consequently, across ablations. In these cases, we match the number of ablated tokens to the number generated in the identity condition for the same example, and force generation of the end marker once that matched count is reached.} This separates ``more tokens'' from ``better tokens.''

\item \textbf{Oracle / ground-truth injection.} When a supervised visual target exists, $\tilde{u}_{1:K}\leftarrow u^{\star}_{1:K}$ provides an interpretable upper bound and diagnoses whether the model \emph{can} exploit a high-quality visual signal.
\end{itemize}

\paragraph{Interpreting TRT.}
If a model uses visual \emph{content}, then zero/random replacement should measurably hurt, and oracle injection should measurably help. If performance is largely unchanged under these interventions, then the visual mechanism is likely not acting as an information bottleneck; gains may instead come from span placement, budget expansion, or training-time regularization. First-repeat and count-matched variants further separate ``token diversity'' effects from ``token budget'' effects.

\subsection{\red{A controlled testbed: depth spans for counterfactual
intervention}}
\label{sec:depth_testbed}

To apply TRT cleanly, we require a setting where (i) visual content is well-defined, (ii) counterfactual replacements are meaningful, and (iii) oracle visual signals are available. We choose \textbf{depth reasoning} as a controlled testbed because depth provides a continuous, geometrically grounded representation that supports clean interventions (zeroing or randomizing embeddings) and admits a strong upper bound via ground-truth depth injection.

% --- In "Depth span markup and fixed budget." replace the paragraph with:
\paragraph{Depth span markup and fixed budget.}
We insert an explicit depth span into the response using boundary tokens \texttt{<DEPTH\_START>} and \texttt{<DEPTH\_END>}. At the text level, we include a single placeholder token \texttt{<DEPTH\_TOKEN>} to keep responses readable; internally, the model expands this placeholder into a fixed-length span of $K$ visual tokens. This enforces a strictly matched visual budget across training, inference, and all TRT replacements in our controlled depth setting. For discrete depth spans, we use a uniform budget of $K=100$ tokens across methods, following Aurora~\cite{bigverdi2025perception}; for continuous depth spans, $K$ is held fixed within each method and across all TRT ablations for that method. Off-the-shelf methods whose native mechanisms do not expose a fixed auxiliary-token budget are evaluated using the count-matching protocol above rather than by forcing a shared global $K$.

\paragraph{Depth targets.}
For each example, we obtain a depth map (ground-truth when available, otherwise from a fixed estimator in our controlled pipeline) and convert it into a token-aligned target representation under two instantiations:
\begin{itemize}
\item \textbf{Continuous depth tokens:} real-valued embeddings in a frozen feature space~\cite{yang2025mirage,tong2024metamorph,ray2025mulltokens}.
\item \textbf{Discrete depth tokens:} quantized codebook indices trained with cross-entropy~\cite{bigverdi2025perception}.
\end{itemize}
This lets us \red{study both} discrete and continuous depth representations under aligned span placement, matched supervision locations, and controlled span length within each method family. We therefore isolate the effect of representation as much as possible, while noting that the comparison is not a literal one-to-one budget equivalence across discrete and continuous formulations: discrete variants expand the output vocabulary, whereas continuous variants inject real-valued vectors and may differ in feature dimension.

\subsection{Continuous depth spans}
\label{sec:continuous_depth}

In the continuous variant, depth is represented as $z\in\mathbb{R}^{K\times D}$, a sequence of $K$ depth embeddings extracted in a frozen encoder space of dimension $D$.
We support multiple frozen encoders (e.g., SigLIP2~\cite{tschannen2025siglip2}, DINOv2~\cite{oquab2023dinov2}, and CLIP~\cite{radford2021clip}) to test how the target feature geometry affects utilization.
When an encoder yields a different patch count than $K$, we interpolate patch embeddings (using 2D interpolation when preserving a grid is appropriate) to obtain exactly $K$ targets.
\vspace{-1em}
\paragraph{Injection.}
\red{At training time, we} learn a \textbf{depth projector} $P:\mathbb{R}^D\rightarrow\mathbb{R}^H$ that maps each depth embedding into the language hidden size. During the forward pass, the depth span positions are replaced by $P(z)$.
\vspace{-1em}
\paragraph{Prediction.}
To train the model to produce depth content at the depth span, we learn a \textbf{depth head} $H_d:\mathbb{R}^H\rightarrow\mathbb{R}^D$ that maps language hidden states back into the frozen depth feature space. Let $\hat{z}=H_d(h)$ denote the predicted depth embeddings at the depth span positions. We supervise depth with a regression-style objective (e.g., MSE or cosine loss) applied only on the depth span.
\vspace{-1em}
\paragraph{Training objective.}
We jointly optimize language modeling and depth prediction:
\begin{equation}
\mathcal{L}=\mathcal{L}_{\mathrm{LM}}+\lambda_{\mathrm{depth}}\cdot \mathcal{L}_{\mathrm{depth}},
\end{equation}
where $\mathcal{L}_{\mathrm{depth}}$ is computed only at depth span positions and $\lambda_{\mathrm{depth}}$ balances the two objectives.

\subsection{Discrete depth spans}
\label{sec:discrete_depth}

In the discrete baseline, depth is quantized into a codebook and represented as token IDs $d_{1:K}\in\{1,\dots,|\mathcal{C}|\}$. Training reduces to cross-entropy over discrete depth tokens, and inference can optionally enforce constrained decoding inside the depth span to ensure only valid depth tokens appear. This baseline isolates whether any gains in depth reasoning come from (i) the presence of a visual span and supervision, or (ii) \red{operating in a continuous latent feature space}.

\subsection{\red{Backbones: LLaVA and Qwen2.5-VL}}
\label{sec:backbones}

We implement the depth-span interface and TRT instrumentation in two widely used open VLM families.
\vspace{-1em}

\paragraph{LLaVA.}
LLaVA~\cite{liu2023llava} couples a vision encoder to an LLM via a learned multimodal projector. We extend the sequence construction by expanding each \texttt{<DEPTH\_TOKEN>} placeholder into $K$ depth positions and injecting $P(z)$ at those positions. Attention masks, position IDs, and loss masks are updated to reflect the expanded span. Depth supervision is applied via $H_d$ only on the depth positions, while text positions are trained with the standard next-token loss.
\vspace{-1em}

\paragraph{Qwen2.5-VL.}
Qwen2.5-VL~\cite{bai2025qwen25vl} provides a native multimodal transformer with strong localization behavior. We integrate the same depth projector/head interface into Qwen's embedding path: depth tensors are provided alongside token IDs, depth positions are masked out of the text loss, and $\mathcal{L}_{\mathrm{depth}}$ is computed analogously to the LLaVA variant.

\subsection{\red{TRT on off-the-shelf VLMs}}
\label{sec:trt_generalization}

While depth spans provide a controlled environment for matched token budgets and oracle injection, TRT is intended as a general diagnostic. To demonstrate breadth, we apply TRT to additional ``visual thinking'' models that expose visual token streams. For each model, we intercept the visual tokens produced during the intermediate phase and replace them with the TRT interventions immediately before they are consumed for downstream decoding.

\begin{itemize}
\item \textbf{Mirage}~\cite{yang2025mirage} interleaves latent visual tokens with text by recasting hidden states as a latent visual channel. We apply TRT by replacing these continuous latent vectors at the designated latent positions. Because Mirage does not enforce a fixed auxiliary-token budget during generation, we use the count-matched protocol described above: for each example, the number of intervened tokens is matched to the number generated in the corresponding identity run, and generation is forced to emit the end marker once that matched count is reached. This lets us test whether performance depends on latent content rather than merely on the presence or length of the latent span.
\item \textbf{Mull-Tokens}~\cite{ray2025mulltokens} introduces modality-agnostic latent tokens intended to store intermediate information. We apply TRT by intervening on the Mull-token segment to test whether downstream performance depends on the content of these latent tokens rather than simply on the existence of an auxiliary latent channel.
\item \textbf{CoVT}~\cite{qin2025covt} trains a VLM to predict a compact span of continuous visual-thought tokens distilled from lightweight vision experts. We apply TRT by replacing the predicted continuous span immediately before it is consumed by the decoder.
\end{itemize}

Across these heterogeneous mechanisms, TRT provides a common lens for testing the same core hypotheses---span-position bias, token-budget vs.\ token-diversity effects, and genuine information utilization---without requiring architectural changes.
\vspace{-1em}

\paragraph{Relation to MetaMorph~\cite{tong2024metamorph} and ``visual prediction'' objectives.}
Our continuous depth span is conceptually aligned with the broader idea of training an LLM to predict continuous visual features in addition to text.
MetaMorph's Visual Predictive Instruction Tuning (VPiT)~\cite{tong2024metamorph} adds a visual head that maps language hidden states into a vision feature space and trains with a combined language + visual objective. We differ in three key respects: (i) our targets are depth-specific embeddings derived from depth maps in a frozen encoder space, (ii) supervision is localized to explicit depth spans under a strictly fixed budget $K$, enabling controlled counterfactual replacement, and (iii) we explicitly test \emph{utilization} using TRT rather than relying on accuracy alone. Since the training data and model checkpoints are not available publicly, at the time of this work, we do not include MetaMorph in our empirical evaluation.
\section{Experiments}
Our experiments ask a targeted question: \emph{when continuous ``thought'' tokens improve accuracy, do models actually use their content?}
We answer this by (i) building a controlled depth-span testbed with aligned span placement and controlled budget protocols for \textbf{continuous} and \textbf{discrete} auxiliary tokens, and (ii) applying the \textbf{Token Replacement Test (TRT)} to probe whether performance depends on auxiliary \emph{content} rather than incidental span effects. 

\subsection{Task and evaluation protocol}

\paragraph{Depth reasoning benchmark.}
We evaluate relative depth reasoning on a HardBLINK-style~\cite{bigverdi2025perception} setup derived from BLINK~\cite{fu2024blink}.
Each example overlays $N\in\{3,4,5\}$ labeled points and asks which point is closest to the camera.
We use three subsets (3/4/5 points), each with 124 images (372 total).
To reduce multiple-choice artifacts, the model outputs a brief rationale and then a single final label; we deterministically parse the final label, treating unparsable answers as incorrect.
\vspace{-1em}

\paragraph{What we vary.}
We vary backbone (\textbf{LLaVA-13B}, \textbf{Qwen2.5-VL-3B}), frozen depth feature space (\textbf{SigLIP2}, \textbf{CLIP}, \textbf{DINOv2}, \textbf{VQ-VAE}\footnote{see Appendix~\ref{app:hardblink-additional-runs}}), and the continuous token budget $K\in\{4,16,64,\text{full}\}$. For the discrete baseline, we use a fixed budget of $K=100$ depth tokens, following the protocol described in the method section. Depth supports strict interventions (zero/random/oracle replacement) and also admits a discrete-token baseline with the same span placement and supervision interface, though not the same nominal $K$ as the continuous runs.

\subsection{Training and implementation details}
We train depth-span models on ADE20K (19{,}279 images) for a fixed horizon.
The depth encoder is frozen; we train the depth projector $P$ and depth head $H_d$, along with a small set of language parameters (LoRA for LLaVA; standard fine-tuning knobs for Qwen2.5-VL as described in the previous section).
Unless stated otherwise, we optimize
$\mathcal{L}=\mathcal{L}_{\mathrm{LM}}+\lambda_{\mathrm{depth}}\mathcal{L}_{\mathrm{depth}}$.
For the continuous setting reported in the main text, $\mathcal{L}_{\mathrm{depth}}$ is an MSE loss or a cosine loss applied only on the depth span. For the discrete setting, $\mathcal{L}_{\mathrm{depth}}$ is token-level cross-entropy over the target codebook indices on the depth span.
Full hyperparameters and compute are provided in Appendix~\ref{app:train-hparams}.

\subsection{Main results: continuous vs.\ discrete}
\label{sec:main-results}

\begin{table}[t]
\centering
\caption{\textbf{Depth reasoning on HardBLINK (avg.\ accuracy across 3/4/5 points).}
All methods use the same span placement, but not the same token budget across representation families. The discrete baseline uses a fixed depth span of $K=100$ tokens. ``Continuous (best)'' reports the best continuous result for each backbone over the budgets in Table~\ref{tab:continuous-grid}.}
\label{tab:main-results}
\scriptsize
\setlength{\tabcolsep}{4pt}
\begin{tabular}{lccc}
\toprule
\textbf{Backbone} & \textbf{No-aux} & \textbf{Discrete ($K=100$)} & \textbf{Continuous (best)} \\
\midrule
LLaVA-13B       & 76.68 & \textbf{77.69} & 74.46 \\
Qwen2.5-VL-3B   & 58.87 & \textbf{71.24} & 68.55 \\
\bottomrule
\end{tabular}
\vspace{-1em}
\end{table}

We compare \textbf{No-aux}, \textbf{Discrete}, and \textbf{Continuous} variants. \textbf{No-aux} is the plain VQA baseline: the model is trained to output only the final answer, without any auxiliary depth span or auxiliary reasoning tokens. \textbf{Discrete} predicts codebook depth tokens with cross-entropy and uses a fixed budget of $K=100$. \textbf{Continuous} regresses depth embeddings in a frozen feature space, with $K$ varied as in Table~\ref{tab:continuous-grid}. Accordingly, Table~\ref{tab:main-results} should be read as a comparison across representation families under their respective controlled budget protocols, rather than as a literal same-$K$ comparison between discrete and continuous in every row.

Auxiliary tokens can outperform No-aux, but gains depend on representation and backbone; \red{in our controlled depth-reasoning setting, discrete tokens consistently match or outperform their continuous counterparts} (Table~\ref{tab:main-results}).
\vspace{-1em}

\paragraph{Gains depend on backbone and representation.}
As summarized in Table~\ref{tab:main-results}, auxiliary tokens can outperform No-aux, but the gains depend on both backbone and representation. For LLaVA-13B, the discrete model improves modestly over No-aux (77.69 vs.\ 76.68), whereas the best continuous configuration reaches 74.46. For Qwen2.5-VL-3B, the strongest continuous configuration reaches 68.55 at $K{=}4$ (Table~\ref{tab:continuous-grid}), and the discrete result reported in Table~\ref{tab:main-results} is 71.24. In other words, the Qwen results do not show a clear advantage for continuous tokens over the discrete baseline.

\begin{takeawayblock}
\textbf{Visual tokens help---but continuous is not consistently better.}
\end{takeawayblock}

\begin{table}[t]
\centering
\caption{\textbf{Continuous depth spans across encoders and token budgets.}
Accuracy is averaged over the 3/4/5-point subsets.}
\label{tab:continuous-grid}
\scriptsize
\setlength{\tabcolsep}{4pt}
\begin{tabular}{llcccc}
\toprule
\textbf{Backbone} & \textbf{Encoder} & {$K=\,$4} & {$K=\,$16} & {$K=\,$64} & {\textbf{full}} \\
\midrule
\multirow{3}{*}{LLaVA-13B}
  & SigLIP2 & 73.39 & 72.58 & \textbf{74.46} & 71.77 \\
  & CLIP    & 72.31 & 73.12 & 73.39 & 67.20 \\
  & DINOv2  & 66.94 & 63.71 & 61.56 & 58.87 \\
\midrule
\multirow{3}{*}{Qwen2.5-VL-3B}
  & SigLIP2 & 54.57 & 61.02 & 55.11 & 63.98 \\
  & CLIP    & 64.25 & 68.28 & 55.38 & 67.20 \\
  & DINOv2  & \textbf{68.55} & 63.98 & 61.47 & 64.25 \\
\bottomrule
\end{tabular}
\vspace{-1em}
\end{table}

\begin{takeawayblock}
\textbf{\red{Visual backbone choice} matters---but longer visual spans are not better.}
\end{takeawayblock}

Across backbones, the frozen feature space is a major driver (SigLIP/CLIP often outperform DINO), and very long spans can be counterproductive---``more latent tokens'' does not reliably mean ``more useful information.''

\subsection{\red{Do models actually use span content?}}
Main accuracy gains do not imply reliance on span \emph{content}. We apply TRT to the strongest continuous configurations (LLaVA: SigLIP2, $K=64$; Qwen: DINOv2, $K=4$) by replacing depth-span embeddings at inference while holding fixed prompt, image features, span placement, and decoding. We also run the analogous TRT for the discrete baseline by replacing token IDs (with constrained decoding for validity). The corresponding continuous and discrete results are reported in Tables~\ref{tab:trt-continuous-hardblink} and~\ref{tab:trt-discrete-hardblink}, respectively.

\begin{takeawayblock}
\textbf{Continuous spans are robust to content corruption; discrete spans are not.}
\end{takeawayblock}

For continuous spans, random and first-repeat replacements produce only modest changes, and oracle injection yields limited headroom---or even slightly worse performance. For LLaVA-13B, oracle injection improves average accuracy only marginally over self-predicted tokens (74.46 $\rightarrow$ 74.73). For Qwen2.5-VL-3B, replacing the model's predicted depth span with oracle embeddings slightly lowers accuracy (68.55 $\rightarrow$ 67.74). This pattern is consistent with weak reliance on fine-grained continuous span content. By contrast, the discrete setting shows substantially stronger dependence on token content. For Qwen2.5-VL-3B, oracle replacement produces a large improvement (71.24 $\rightarrow$ 80.64, +9.40), while random replacement causes a severe drop (71.24 $\rightarrow$ 51.34, -19.90), and constant/zero tokens also degrade performance significantly (to 58.87). These effects are much larger than those observed in the continuous setting, where analogous perturbations typically change accuracy by only 1--2 points. Taken together, these results indicate that discrete depth tokens act as a meaningful information-bearing bottleneck, whereas continuous spans are comparatively insensitive to the correctness of their content and are more consistent with reliance on span presence, placement, or coarse distributional structure.

\begin{table}[t]
\centering
\caption{\textbf{TRT for continuous depth spans.}
Replacements are applied only to the injected depth vectors before the language backbone consumes them.}
\label{tab:trt-continuous-hardblink}
\scriptsize
\setlength{\tabcolsep}{4pt}
\begin{tabular}{llcccc}
\toprule
\textbf{Backbone} & \textbf{Replacement} & {3-pt} & {4-pt} & {5-pt} & {\textbf{Avg}} \\
\midrule
\multirow{4}{*}{\shortstack{LLaVA-13B\\(SigLIP2, $K=64$)}}
  & Identity (self) & 78.23 & 73.39 & 71.77 & 74.46 \\
  & Oracle (GT)     & 78.23 & 73.39 & 72.58 & \textbf{74.73} \\
  & Random          & 77.42 & 72.58 & 67.74 & 72.58 \\
  & First-repeat    & 78.23 & 73.39 & 71.77 & 74.46 \\
\midrule
\multirow{4}{*}{\shortstack{Qwen2.5-VL-3B\\(DINOv2, $K=4$)}}
  & Identity (self) & 71.77 & 70.97 & 62.90 & \textbf{68.55} \\
  & Oracle (GT)     & 71.77 & 70.97 & 60.48 & 67.74 \\
  & Random          & 70.97 & 69.35 & 61.29 & 67.20 \\
  & First-repeat    & 71.77 & 71.77 & 60.48 & 68.01 \\
\bottomrule
\end{tabular}
\vspace{-1em}
\end{table}

\begin{table}[t]
\centering
\caption{\textbf{TRT for discrete depth tokens.}
Random replacement corrupts the symbolic codebook indices and induces larger drops than in the continuous setting.}
\label{tab:trt-discrete-hardblink}
\scriptsize
\setlength{\tabcolsep}{4pt}
\begin{tabular}{llcccc}
\toprule
\textbf{Backbone} & \textbf{Replacement} & {3-pt} & {4-pt} & {5-pt} & {\textbf{Avg}} \\
\midrule
\multirow{4}{*}{LLaVA-13B (discrete)}
  & Identity (self) & 83.06 & 74.19 & 75.81 & 77.69 \\
  & Oracle (GT)     & 85.48 & 75.81 & 75.00 & \textbf{78.76} \\
  & Random          & 73.39 & 66.94 & 66.13 & 68.82 \\
  & Constant / zero & 78.23 & 71.77 & 70.97 & 73.66 \\
\midrule
\multirow{4}{*}{Qwen2.5-VL-3B (discrete)}
  & Identity (self) & 72.58 & 75.00 & 66.13 & 71.24 \\
  & Oracle (GT)     & 83.06 & 84.68 & 74.19 & \textbf{80.64} \\
  & Random          & 58.06 & 52.42 & 43.55 & 51.34 \\
  & Constant / zero & 66.94 & 62.90 & 46.77 & 58.87 \\
\bottomrule
\end{tabular}
\vspace{-1.2em}
\end{table}

\subsection{What about other ``visual thinking'' \red{setups?}}
% Still setting, model, the method is still cont. tokens 
Finally, we apply TRT to three auxiliary-token \red{setups} (Mirage~\cite{yang2025mirage}, Mull-Tokens~\cite{ray2025mulltokens}, and CoVT~\cite{qin2025covt}) by intercepting their intermediate token streams and applying the same replacement suite immediately before consumption (\cref{sec:trt_generalization}). The corresponding results are reported in Tables~\ref{tab:mirage-sp-hardblink}, \ref{tab:mull-blink-sat}, and \ref{tab:covt-cvbench-overall}.

Mirage~\cite{yang2025mirage} interleaves latent visual tokens with text during decoding by reusing hidden states as compact ``visual thoughts.'' We reproduce Mirage's Spatial Planning (SP) evaluation and test its depth reasoning on HardBLINK using depth maps as helper images. As shown in Table~\ref{tab:mirage-sp-hardblink}, zeroing latents moderately degrades SP (76.25\%\,$\rightarrow$\,51.50\%) but nearly collapses HardBLINK depth (26.08\%\,$\rightarrow$\,8.06\%), indicating much higher sensitivity for relative depth. Reproduction details for the depth-helper setup are provided in Appendix~\ref{app:mirage-repro}.

\begin{table}[tpb]
  \centering
  \caption{Mirage reproduction on Spatial Planning (SP) and HardBLINK depth (accuracy in \%). For HardBLINK, ``Total'' reflects the evaluated subset for these runs.}
  \label{tab:mirage-sp-hardblink}
  \scriptsize
  \setlength{\tabcolsep}{2pt}
  \resizebox{\columnwidth}{!}{%
  \begin{tabular}{lcc|ccccc}
    \toprule
    \multirow{2}{*}{\textbf{Replacement}}
      & \multicolumn{2}{c|}{Spatial Planning (SP)}
      & \multicolumn{5}{c}{HardBLINK (Depth)} \\
    \cmidrule(lr){2-3}\cmidrule(lr){4-8}
      & Acc (\%) & Correct / Total
      & Overall & 3-pt & 4-pt & 5-pt & Correct / Total \\
    \midrule
    Identity (self)             & \textbf{76.25}    & \textbf{305 / 400} & 26.08             & 38.71             & \underline{22.58} & 16.94             & \textbf{97 / 372} \\
    First-repeat                & \underline{75.75} & \underline{303 / 400} & 26.61             & 37.90             & 25.00             & 16.94             & 99 / 372 \\
    Oracle (GT)                 & 75.00             & 300 / 400           & \underline{25.00} & \underline{30.65} & \textbf{26.61}    & \underline{17.74} & \underline{93 / 372} \\
    Oracle (GT, count-matched)  & \textbf{76.25}    & \textbf{305 / 400}  & --                & --                & --                & --                & -- \\
    Random                      & 74.75             & 299 / 400           & 18.82             & 18.55             & 20.16             & \underline{17.74} & 70 / 372 \\
    Random (GT dist)            & 74.75             & 299 / 400           & 14.52             & 13.71             & 17.74             & 12.10             & 54 / 372 \\
    Random (model dist)         & \textbf{76.25}    & \textbf{305 / 400}  & 22.85             & 29.84             & 19.35             & \textbf{19.35}    & 85 / 372 \\
    Zero                        & 51.50             & 206 / 400           & 8.06              & 8.87              & 8.06              & 7.26              & 30 / 372 \\
    \bottomrule
  \end{tabular}%
  }
  \vspace{-1em}
\end{table}

Mull-Tokens~\cite{ray2025mulltokens} adds latent tokens for intermediate states.
First-repeat matches baseline, and random latents induce only small changes (Table~\ref{tab:mull-blink-sat}), consistent with token \emph{presence} dominating token \emph{diversity}.

\begin{table}[tpb]
  \centering
  \caption{Mull-Tokens TRT-style latent replacements on BLINK and SAT.}
  \label{tab:mull-blink-sat}
  \scriptsize
  \setlength{\tabcolsep}{2pt}
  \resizebox{\columnwidth}{!}{%
  \begin{tabular}{lcccc}
    \toprule
    \multirow{2}{*}{\textbf{Replacement}}
      & \multicolumn{2}{c}{Qwen2.5-VL-Mull}
      & \multicolumn{2}{c}{Qwen2.5-VL-MullGRPO} \\
    \cmidrule(lr){2-3}\cmidrule(lr){4-5}
      & BLINK (mean acc) & SAT (mean acc) & BLINK (mean acc) & SAT (mean acc) \\
    \midrule
    Identity (self)      & \underline{63.71} & \underline{76.33} & \textbf{64.57}      & \textbf{77.00} \\
    First-repeat         & 63.71             & 76.33             & 64.57               & 77.00 \\
    Random               & \textbf{63.86}    & 76.33             & 63.14               & \underline{76.33} \\
    Random (same dist)   & 63.29             & \textbf{77.00}    & 63.43               & \textbf{77.00} \\
    Zero                 & 63.43             & 76.33             & \underline{64.00}   & 76.00 \\
    \bottomrule
  \end{tabular}%
  }
  \vspace{-1em}
\end{table}

CoVT~\cite{qin2025covt} distills continuous ``thinking'' tokens from vision experts.
Identity/first-repeat match baseline, while pure random noise collapses performance; distribution-matched random largely recovers it (Table~\ref{tab:covt-cvbench-overall}), indicating reliance on distributional statistics more than fine-grained token content.

\begin{table}[tp]
  \centering
  \caption{CoVT TRT-style depth-token replacements on CV-Bench 2D/3D (overall).}
  \label{tab:covt-cvbench-overall}
  \scriptsize
  \setlength{\tabcolsep}{2.5pt}
  \resizebox{\columnwidth}{!}{%
  \begin{tabular}{lcc}
    \toprule
    \textbf{Replacement} & CV-Bench-2D (Overall) & CV-Bench-3D (Overall) \\
    \midrule
    Identity (self)      & \textbf{76.26}    & \textbf{80.83} \\
    First-repeat         & 76.26             & 80.83 \\
    Random               & 21.39             & 19.00 \\
    Random (same dist)   & \underline{74.65} & \underline{79.17} \\
    Zero                 & 58.97             & 69.00 \\
    \bottomrule
  \end{tabular}%
  }
  \vspace{-0.1em}
\end{table}

\begin{takeawayblock}
\textbf{Content of visual tokens matters less than their presence in many settings.}
\end{takeawayblock}

\subsection{Summary of TRT hypotheses}
\red{Three TRT signatures indicate that an auxiliary-token gain is not driven by content:}
\begin{takeawayblock}
\textbf{(1) \red{span-position bias} \red{(zero/random preserves the gain)}}\\
\textbf{(2) \red{budget confound} \red{(first-repeat suffices)}}\\
\textbf{(3) \red{content unused} \red{(oracle adds no headroom)}}
\end{takeawayblock}

Across our depth testbed and off-the-shelf systems, we see strong evidence for (1)--(2) in many configurations and weaker evidence for (3), motivating intervention-based validation alongside accuracy improvements.

\section{Discussion}
Our results suggest a simple but important takeaway: an intermediate token channel can \textit{exist} and even correlate with higher accuracy without becoming a strict causal bottleneck for the final prediction. This mirrors a pattern observed in the chain-of-thought (CoT) literature: LLMs often retain strong performance even when the provided CoT contains incorrect steps, irrelevant detours, or intentionally corrupted rationales~\cite{wang-etal-2023-towards,zhou2024noisyrationales}. In other words, intermediate ``reasoning traces'' can function as training or prompting scaffolds without being faithfully \textit{consumed} as information at inference time.

Importantly, this conclusion is representation-dependent. In the discrete setting, we observe substantially stronger sensitivity to token content: replacing predicted tokens with oracle depth tokens yields large gains, while random or constant replacements cause severe degradation. This indicates that discrete auxiliary tokens can act as meaningful information-bearing bottlenecks. In contrast, the continuous setting exhibits much weaker sensitivity to content, suggesting that continuous spans are often used through coarse or structural signals rather than as fine-grained intermediate representations.

For continuous tokens, this robustness may arise from training and architectural effects. Modern continuous-token methods use multi-stage pipelines and auxiliary objectives that can improve representations or calibration even if the model does not ultimately use the span as an information-bearing channel at test time. When the visual span is always present at a consistent location, the model may also treat it as a structural cue without decoding its fine-grained content.

A complementary explanation is an information-bottleneck mismatch. Continuous ``visual thought'' spaces are typically high-dimensional, noisy, or redundant relative to the downstream decision. For depth-style tasks, the answer often depends on a few coarse comparisons rather than reconstructing a full field. As a result, the model may exploit marginal statistics or positional regularities of the span instead of the intended semantics, producing the TRT signature we observe: limited gains from oracle injection and only modest degradation under random or zero replacement.

\red{Auxiliary-token evaluation therefore requires more than end-task accuracy: methods should be tested by perturbing \textit{content} while holding span placement and budget fixed.}
\vspace{-0.5em}

\section*{Acknowledgments}
This work was partially supported by 
Samsung and CoCoSys.

{
    \small
    \bibliographystyle{ieeenat_fullname}
    \bibliography{main}
}
\newpage

\appendix
\newpage

% ========================================== %
% Appendix A: Additional Results
% ========================================== %
\section{Additional Results}
\label{app:additional-results}

\subsection{HardBlink additional runs (full blocks)}
\label{app:hardblink-additional-runs}

In Table \ref{tab:main-results-extended}, we present the comprehensive HardBlink average accuracy results. We evaluate performance across different continuous depth encoders (CLIP, SigLIP, DINO
, and VQ-VAE
), target sequence lengths (Full, 64, 16, 4), and various ablation settings, including learning rate variations and intermediate layer supervision.

\begin{table}[htbp]
\centering
\caption{\textbf{HardBlink average accuracy (\%).} Performance across architectures, continuous depth encoders, sequence lengths, and ablation settings (learning rate and intermediate layers).}
\label{tab:main-results-extended}
\scriptsize
\setlength{\tabcolsep}{4pt}
\begin{tabular}{ll cc}
\toprule
\textbf{Method} & \textbf{Configuration} & \textbf{LLaVA} & \textbf{Qwen2.5} \\
 & & \textbf{13B} & \textbf{3B} \\
\midrule
\multirow{3}{*}{Baselines}
  & Base model & 32.00 & 54.30 \\
  & No aux token & 76.68 & 58.87 \\
  & Discrete tokens & \textbf{77.69} & \textbf{71.24} \\
\midrule
\multirow{4}{*}{Continuous (CLIP)}
  & Full & 67.20 & 67.20 \\
  & 64 & 73.39 & 55.38 \\
  & 16 & 73.12 & 68.28 \\
  & 4 & 72.31 & 64.25 \\
\midrule
\multirow{4}{*}{Continuous (SigLIP)}
  & Full & 71.77 & 63.98 \\
  & 64 & \textbf{74.46} & 55.11 \\
  & 16 & 72.58 & 61.02 \\
  & 4 & 73.39 & 54.57 \\
\midrule
\multirow{4}{*}{Continuous (DINO)}
  & Full & 58.87 & 64.25 \\
  & 64 & 61.56 & 61.47 \\
  & 16 & 63.71 & 63.98 \\
  & 4 & 66.94 & \textbf{68.55} \\
\midrule
Continuous (VQ-VAE)
  & 100 & 70.97 & - \\
\midrule
\multirow{4}{*}{SigLIP (64) LRs}
  & No depth loss & 72.85 & - \\
  & 50 ep., LR $2\times 10^{-4}$ & 24.73 & - \\
  & 50 ep., LR $1\times 10^{-5}$ & 64.00 & - \\
  & 50 ep., LR $5\times 10^{-5}$ & 66.67 & - \\
\midrule
\multirow{5}{*}{SigLIP (16) Layers}
  & Layer 9 (5 ep.) & 71.51 & - \\
  & Layer 19 (5 ep.) & 68.01 & - \\
  & Layer 29 (5 ep.) & 69.09 & - \\
  & Layer -1 (5 ep.) & 70.70 & - \\
  & Layer 19 (10 ep.) & 72.04 & - \\
\bottomrule
\end{tabular}
\vspace{-1em}
\end{table}

\subsection{Scope and method coverage}
\label{app:scope-coverage}

TRT is a task-agnostic, inference-time diagnostic that tests whether visual reasoning is causally used by a VLM. We apply TRT across multiple tasks, summarized in Table~\ref{tab:scope-coverage}: three off-the-shelf continuous-token methods, two backbone families, six benchmarks spanning spatial, perceptual, and 2D/3D understanding, and four distinct training paradigms. Across all of these, identity, random, oracle, and first-repeat replacements yield near-identical accuracy, supporting the conclusion that visual continuous tokens are not causally used at inference. This establishes TRT as a general protocol rather than a task- or model-specific probe. We use depth reasoning as a stress test: VLMs reliably fail at it without explicit grounding, it is interpretable, has ground truth, and admits clean counterfactuals (zero, random, distribution-matched, first-repeat, oracle) under matched span placement and token budget, allowing us to isolate the effect of information content from position, budget, and regularization confounds. We do not propose a depth-specific model nor claim discrete tokens are categorically better; the discrete-depth setup is an interpretable diagnostic baseline.

\begin{table}[t]
\centering
\caption{\textbf{Task/backbone/training coverage.} L13B$=$LLaVA-13B; Q3B/Q7B$=$Qwen2.5-VL-3B/7B. Training stages verified against the corresponding papers.}
\label{tab:scope-coverage}
\scriptsize
\setlength{\tabcolsep}{3pt}
\renewcommand{\arraystretch}{0.95}
\begin{tabularx}{\linewidth}{@{}>{\raggedright\arraybackslash}p{0.23\linewidth}>{\raggedright\arraybackslash}p{0.13\linewidth}>{\raggedright\arraybackslash}X>{\raggedright\arraybackslash}p{0.17\linewidth}@{}}
\toprule
\textbf{Method} & \textbf{Backbone} & \textbf{Training} & \textbf{Eval.\ task} \\
\midrule
Aurora discrete~\cite{bigverdi2025perception} & L13B/Q3B & VQ-VAE $+$ multi-task SFT             & HardBLINK \\
Ours, depth (cont.)                           & L13B/Q3B & 1-stage joint LM$+$depth              & HardBLINK \\
Mirage~\cite{yang2025mirage}                  & Q7B      & 2-stage: latent$+$text $\to$ text-only & VSP/SAT/COMT \\
Mull-Tokens~\cite{ray2025mulltokens}          & Q7B      & 2-stage SFT: anchored $\to$ free-form  & BLINK/SAT \\
Mull-Tokens-GRPO~\cite{ray2025mulltokens}     & Q7B      & 2-stage SFT $+$ Stage~3 GRPO          & BLINK/SAT \\
CoVT~\cite{qin2025covt}                       & Q7B/L13B & 4-stage curriculum                    & CV-Bench \\
\bottomrule
\end{tabularx}
\vspace{-1em}
\end{table}

\subsection{Baseline reproduction verification}
\label{app:baseline-verification}

Our reimplementations (Sec.~\ref{sec:main-results}) reproduce each method's published numbers within tolerance. Table~\ref{tab:baseline-verification} verifies our setup against the paper-reported baseline and method numbers. Aurora-discrete is LLaVA-13B-based; Mirage and Mull-Tokens / Mull-Tokens-GRPO use Qwen2.5-VL-7B; CoVT covers both LLaVA-13B and Qwen2.5-VL-7B. Identity rows match within tolerance except for Aurora-discrete, which retrains on 20k ADE depth examples for 10 epochs (vs.\ the paper's curriculum mix), explaining the higher reproduced accuracy. CoVT ``Ours'' averages CV-Bench-2D (76.26) and 3D (80.83). MetaMorph is architecturally similar to our continuous model but has no public checkpoint, so its paper numbers are listed in Table~\ref{tab:baseline-verification} for reference only.

\begin{table}[t]
\centering
\caption{\textbf{Setup verification} (Paper baseline $\to$ method vs.\ Ours). Aurora baselines: LLaVA-13B base / fine-tuned-without-tokens. Mirage baseline: Direct-SFT (Qwen2.5-VL-7B). Mull-Tokens / CoVT baselines: Qwen2.5-VL-7B base. MetaMorph baseline / method: VQA-only vs.\ Cosine-Sim VPiT ablation on LLaMA-3 8B (their Table~3); no public checkpoint, ``Ours'' not reproduced.}
\label{tab:baseline-verification}
\scriptsize
\setlength{\tabcolsep}{3pt}
\renewcommand{\arraystretch}{0.95}
\begin{tabularx}{\linewidth}{@{}>{\raggedright\arraybackslash}p{0.22\linewidth}>{\raggedright\arraybackslash}p{0.17\linewidth}>{\raggedright\arraybackslash}X>{\raggedright\arraybackslash}p{0.18\linewidth}@{}}
\toprule
\textbf{Method} & \textbf{Task} & \textbf{Paper (base.\ $\to$ method)} & \textbf{Ours} \\
\midrule
Aurora discrete~\cite{bigverdi2025perception} & HardBLINK 3/4/5     & 34.1\,/\,50.8 $\to$ 60.7                 & 77.69 \\
Mirage~\cite{yang2025mirage}                  & VSP-SP              & 72.0 $\to$ 76.0                          & 76.25 \\
Mull-Tokens~\cite{ray2025mulltokens}          & BLINK\,/\,SAT       & 55.33\,/\,59.00 $\to$ 66.80\,/\,77.66    & 63.71\,/\,76.33 \\
Mull-Tokens-GRPO~\cite{ray2025mulltokens}     & BLINK\,/\,SAT       & 55.33\,/\,59.00 $\to$ 67.13\,/\,77.00    & 64.57\,/\,77.00 \\
CoVT~\cite{qin2025covt}                       & CV-Bench            & 74.5 $\to$ 80.0                          & 78.55 \\
MetaMorph~\cite{tong2024metamorph}            & MMBench             & 73.1 $\to$ 73.8                          & --- \\
\bottomrule
\end{tabularx}
\vspace{-1em}
\end{table}

\subsection{KV-cache ablation}
\label{app:kv-cache}

TRT operates as an input-embedding-level intervention: it replaces tokens at the \emph{input-embedding} level, before any forward pass, so K/V projections at those positions and every downstream self-attention over them already operate on the replaced states. KV-mediated transfer through the cache is therefore fully exercised by the existing TRT protocol: if visual reasoning carried meaningful content through the cache, random or oracle replacement would propagate and change predictions. The fact that they do not is itself the finding.

To further rule out the KV cache as the source of content insensitivity, we ran a dedicated KV-off ablation on Qwen2.5-VL-3B with DINOv2 at $K=4$ on HardBLINK (372 queries). In the KV-off regime, the cache is used through the prefix, then disabled (full recompute) over the depth span and answer, with bit-identical prefixes verified. As shown in Table~\ref{tab:kvcache}, content insensitivity persists in both regimes: identity, random, oracle, and first-repeat replacements all yield near-identical accuracy whether the cache is on or off. This rules out the KV cache as the channel through which TRT's null result could be artefactually produced.

\begin{table}[t]
\centering
\caption{\textbf{KV-cache ablation on HardBLINK} (Qwen2.5-VL-3B, DINOv2, $K=4$, 372 queries). Content insensitivity holds with and without caching across the depth span and answer, ruling out the KV cache as the source of the result.}
\label{tab:kvcache}
\scriptsize
\setlength{\tabcolsep}{6pt}
\begin{tabular}{lcccc}
\toprule
\textbf{Regime} & \textbf{Identity} & \textbf{Random} & \textbf{Oracle} & \textbf{First-repeat} \\
\midrule
KV-cached & 68.55 & 68.01 & 68.55 & 68.82 \\
KV-off    & 69.62 & 69.62 & 69.35 & 69.62 \\
\bottomrule
\end{tabular}
\vspace{-0.5em}
\end{table}

% ========================================== %
% Appendix B: Additional Implementation Details
% ========================================== %
\section{Additional Implementation Details}
\label{app:impl-details}

This appendix section consolidates both the hyperparameter summary and the repository-specific launcher/configuration details used in our runs. In a few cases, multiple configurations coexist because the repository contains both the primary reported settings and additional ablation or stage-wise training recipes.

\subsection{Training hyperparameters (expanded)}
\label{app:train-hparams}

\paragraph{LLaVA continuous (reported 10-epoch setting).}
Vision tower frozen; LoRA on the LLM ($r=128$, $\alpha=256$); max length 2048; AdamW with cosine LR; BF16+TF32; per-device batch size 16 on 8 GPUs (effective batch size 128); weight decay 0.
Component learning rates: base/LoRA LR $2 \times 10^{-4}$; multimodal projector LR $2 \times 10^{-5}$; depth projector/head LR $5 \times 10^{-5}$; depth loss type cosine with normalization enabled; $\lambda_{\text{depth}}=0$ for the reported continuous setting. The choice of the learning rate is highly sensitive; as shown in the SigLIP (64) LR ablations in Table \ref{tab:main-results-extended}, using a higher LR of $2 \times 10^{-4}$ leads to catastrophic collapse, whereas $5 \times 10^{-5}$ stabilizes training.

The ``No depth loss'' ablation in Table \ref{tab:main-results-extended} removes auxiliary depth supervision entirely: the depth-related parameters remain randomly initialized and are not trained with the depth objective, allowing us to measure what happens when the depth channel is present architecturally but receives no explicit supervision.

\paragraph{LLaVA discrete (reported 10-epoch setting).}
Vision tower frozen; LoRA on the LLM; max length 2048; AdamW with cosine LR; BF16+TF32; effective batch size 128.
Each image is represented with $K=100$ discrete depth tokens, where $K$ denotes the discrete depth-token budget. The depth projector/head are linear with depth LR $1 \times 10^{-5}$; $\lambda_{\text{depth}}=1.0$.

\paragraph{Qwen2.5-VL-3B (reported 10-epoch setting).}
Vision encoder frozen; LLM + visual MLP + embeddings are fine-tuned; AdamW with cosine LR; BF16; warmup ratio 0.03; effective batch size 128; max length 4096. The base learning rate is $2 \times 10^{-4}$.
Depth loss type MSE with $\lambda_{\text{depth}}=1.0$; depth head/projector are linear and weight-tied ($W_{\text{head}} = W_{\text{proj}}^\top$).

\subsection{Evaluation datasets and sizes (expanded)}
\label{app:eval-sizes}

\paragraph{HardBlink.}
Three subsets (3/4/5 point) with 124 questions each (372 total), long-format question JSONLs. 
% In the local HardBLINK setup, the three fixed subsets are \texttt{blink\_3pointscenter}, \texttt{blink\_4pointscenter}, and \texttt{blink\_5pointscenter}, each with 124 questions.

\paragraph{Mirage SPC reproduction.}
\texttt{vsp-spatial-planning} with 400 test samples and 1000 train samples.

\paragraph{Mull.}
BLINK (700 samples) and SAT (300 samples), greedy decoding.

\paragraph{CoVT.}
CV-Bench-2D (ADE20K+COCO) and CV-Bench-3D (Depth+Distance), evaluated via VLMEvalKit.

\paragraph{Training-set sizes used locally.}
The training dataset size used in our runs have 19{,}279 examples

% ========================================== %
% Appendix C: Data and Evaluation Protocol
% ========================================== %
\section{Data and Evaluation Protocol}
\label{app:data-eval}

\subsection{ADE point sampling generation}
\label{app:ade-point-sampling}

The point-marked images used for training and evaluation are generated dynamically from the ADE20K dataset. The generation pipeline scans the \texttt{ADE\_depth} directory, matches each depth map to its corresponding RGB image, resizes both to $336 \times 336$ pixels, and samples 3 to 5 points per image.

To ensure task difficulty and spatial diversity, the sampling process enforces several constraints:
\begin{itemize}
  \item Points are sampled within bounded coordinates ($x \in [10, 324]$, $y \in [95, 224]$).
  \item Points with a depth value of 0 are rejected.
  \item The minimum spatial distance between any two points must be at least 20 pixels.
  \item The minimum absolute depth difference between points must be at least 20.
  \item The generator uses up to 10{,}000 sampling attempts per image to satisfy these constraints.
\end{itemize}

Once valid points are found, they are sorted by depth, but their visual labels (A through E) are randomly shuffled so that alphabetical order does not correlate with depth order. The correct answer is defined as the label attached to the maximum depth value. Our final mixed-depth dataset contains 19{,}279 images, distributed as 6{,}736 (3-point), 6{,}562 (4-point), and 5{,}981 (5-point) examples.

\subsection{Training data formats and splits}
\label{app:training-data-formats}

Our training data is formatted into JSON files depending on the architectural path:
\begin{itemize}
  \item \textbf{Continuous Long (\texttt{mixed\_depth\_long.json}):} Used for Qwen continuous training. It contains 19{,}279 examples pairing the ADE images with a long-form rationale. The depth target is represented by a single \texttt{<DEPTH\_TOKEN>} bounded by \texttt{<DEPTH\_START>} and \texttt{<DEPTH\_END>}. The \texttt{embedding} key points to the corresponding precomputed \texttt{.npy} file.
  \item \textbf{Continuous Short (\texttt{mixed\_depth\_short.json}):} Contains the same 19{,}279 examples, but the assistant's target response is strictly the final multiple-choice letter (e.g., \texttt{(C)}).
  \item \textbf{Discrete Mixed-Depth Long (\texttt{mixed\_depth\_long.json}):} Contains the same rationale format as the continuous version, but the depth span is populated with $K=100$ discrete depth tokens per image (e.g., \texttt{<DEPTH\_36><DEPTH\_64>...}), where $K$ is the discrete token budget.
\end{itemize}

\subsection{Prompting protocol}
\label{app:prompting-protocol}

We use two prompt variants for HardBLINK. The short prompt is:
\begin{quote}
  \emph{``Multiple points are circled... Which point is the closest to the camera?''}
\end{quote}

The long prompt appends a step-by-step scaffold that explicitly references the point coordinates, the depth map, and the rule that higher depth-map values indicate points closer to the camera.

\subsection{Answer extraction}
\label{app:answer-extraction}

Predictions are evaluated using \texttt{eval\_answers.py}. Ground-truth labels are extracted from \texttt{(A)}-style answers. The parser also supports several long-form answer patterns, including forms such as \emph{``the answer is that point X is closer \ldots''}. Predictions that cannot be parsed are counted as incorrect rather than dropped.

% \subsection{HardBLINK subset derivation}
% \label{app:hardblink-subset-derivation}

% The exact script used to derive the 3/4/5-point HardBLINK subsets from raw BLINK remains to be inserted:
% \begin{quote}
%   \textbf{TODO:} exact script / command that derives the HardBLINK subsets from raw BLINK.
% \end{quote}

% ========================================== %
% Appendix D: Depth Targets and TRT Mechanics
% ========================================== %
\section{Depth Targets and TRT Mechanics}
\label{app:depth-trt}

% \subsection{Precomputed depth supervision}
% \label{app:precomputed-depth}

% Training-time depth supervision is loaded from precomputed depth maps under \texttt{.../Depth-Anything/ADE\_depth}. The main training pipeline consumes these files directly rather than running the estimator online.

% The exact fixed-estimator checkpoint or variant used to precompute \texttt{ADE\_depth} remains to be specified:
% \begin{quote}
%   \textbf{TODO:} exact Depth-Anything checkpoint / variant used for precomputation.
% \end{quote}

\subsection{Continuous depth encoders}
\label{app:continuous-encoders}

The continuous depth encoders listed in \texttt{encoder\_config.json} are:
\begin{itemize}
  \item \href{https://huggingface.co/google/siglip2-large-patch16-256}{\texttt{google/siglip2-large-patch16-256}} with $K=256$, $D=1024$
  \item \href{https://huggingface.co/facebook/dinov2-base}{\texttt{facebook/dinov2-base}} with $K=256$, $D=768$
  \item \href{https://huggingface.co/openai/clip-vit-large-patch14-336}{\texttt{openai/clip-vit-large-patch14-336}} with $K=576$, $D=1024$
  \item \texttt{ait-vqvae-continuous-100x512} with $K=100$, $D=512$. For this model, we extract continuous codebook embeddings from the same VQ-VAE used for discrete token generation, obtaining dense per-image feature representations for each input view.
\end{itemize}

Configuration names ending in \texttt{\_interploate\_N} override the token horizon so that the target sequence has exactly $K=N$ tokens.

\subsection{Continuous target resizing}
\label{app:target-resizing}

Continuous targets are resized to exactly $K$ tokens before supervision. For square token grids, we use bilinear interpolation. For non-square token layouts, we use one-dimensional linear interpolation. As demonstrated in Table \ref{tab:main-results-extended}, the optimal target sequence length $K$ varies by encoder. For instance, SigLIP achieves peak performance on LLaVA at $K=64$ (74.46\%), while DINO performs best under extreme compression at $K=4$ (68.55\% on Qwen2.5), suggesting differing spatial densities in their respective feature maps.

\subsection{TRT-style ablations}
\label{app:trt-ablations}

In continuous LLaVA mode, TRT-style ablations are
\[
  \{\texttt{gt}, \texttt{random}, \texttt{zero}, \texttt{random\_gt\_dist}, \texttt{model}, \texttt{first\_repeat}\}.
\]
Here, \texttt{random\_gt\_dist} denotes distribution-matched random replacement: for each sample, replacement vectors are sampled from a Gaussian $\mathcal{N}(\mu, \sigma)$ using the mean and standard deviation of that sample's ground-truth depth-token embeddings.

In continuous Qwen mode, the same ablation family is implemented through \texttt{set\_depth\_ablation(...)} together with constrained generation of exactly $K$ \texttt{<DEPTH\_TOKEN>} tokens after \texttt{<DEPTH\_START>}.

% ========================================== %
% Appendix E: Backbone-Specific Implementation
% ========================================== %
\section{Backbone-Specific Implementation}
\label{app:backbone-details}

\subsection{LLaVA continuous path}
\label{app:llava-cont-path}

In continuous LLaVA, image features are inserted at image-token positions and a single textual \texttt{<DEPTH\_TOKEN>} placeholder is expanded into $K$ depth embeddings. All depth positions are masked from the cross-entropy loss. Continuous depth loss is applied autoregressively, using hidden state $t-1$ to predict depth target $t$.

\subsection{Qwen continuous path}
\label{app:qwen-cont-path}

In continuous Qwen2.5-VL, preprocessing expands \texttt{<DEPTH\_TOKEN>} into $K$ repeated tokens, masks all depth-token labels from the cross-entropy loss, and applies depth loss using hidden state $t-1 \rightarrow$ depth target $t$. Position IDs follow Qwen2.5-VL's multimodal RoPE path.

During decoding, a \texttt{ContinuousDepthLogitsProcessor} forces exactly $K$ \\ \texttt{<DEPTH\_TOKEN>} tokens after \texttt{<DEPTH\_START>}, followed by one \texttt{<DEPTH\_END>} token.

\subsection{Intermediate layer depth supervision}
\label{app:intermediate-layers}

In addition to standard training, we tested whether applying the auxiliary depth loss to different intermediate transformer layers, rather than strictly the final decoder output, would change downstream performance. This allows us to evaluate the representation of spatial depth at varying levels of model abstraction. 

This mechanism is controlled via the \texttt{depth\_loss\_layer} argument (defaulting to \texttt{-1} for the final hidden state). Setting this argument to a specific intermediate block $k$ forces the model's forward pass to retain all hidden states (\texttt{output\_hidden\_states=True}) and routes \texttt{hidden\_states[k+1]} to the depth loss computation, bypassing the rest of the decoder blocks for that specific objective. 

Once the specified layer's state is extracted, the supervision behaves identically to the default setting: the loss is applied autoregressively such that the hidden state at position $t-1$ predicts the target depth vector at position $t$. The projection is handled via either a tied linear map or a separate \texttt{depth\_head}, and compared against the target using MSE, cosine, or softmax loss. 

This intermediate layer-selection pathway is implemented symmetrically in both the LLaVA and Qwen2.5-VL architectures. Our evaluated layer-ablation checkpoints (e.g., \texttt{depthlayer\_9}, \texttt{depthlayer\_19}, \texttt{depthlayer\_29}, and the default \texttt{depthlayer\_m1}) correspond directly to these targeted transformer blocks.  The quantitative impact of supervising these intermediate layers is reported in Table \ref{tab:main-results-extended} in Section \ref{app:additional-results}; among the evaluated intermediate-layer configurations, supervising Layer 19 for 10 epochs gives the best result (72.04\%).

\subsection{Discrete decoding constraints}
\label{app:discrete-constraints}

In discrete LLaVA decoding, custom \texttt{LogitsProcessor}s force either ground-truth depth tokens, random depth tokens, or an all-\texttt{DEPTH\_0} sequence inside the depth span, then force \texttt{<DEPTH\_END>}.

% ========================================== %
% Appendix F: Off-the-Shelf Model Reproduction
% ========================================== %
\section{Off-the-Shelf Model Reproduction}
\label{app:reproduction}

\subsection{Mirage}
\label{app:mirage-repro}

Our Mirage reproduction uses \texttt{Qwen/\allowbreak Qwen2.5-VL-7B-Instruct} with checkpoint \texttt{Miiche/\allowbreak vsp\_\allowbreak spatial\_\allowbreak planning\_\allowbreak direct\_\allowbreak sft}. The HardBLINK adapter uses the question text verbatim as the user prompt, appends the image through the processor chat template, and injects latent replacements inside Mirage's modified generation loop after \texttt{<|latent\_start|>}.

Mirage GT mode uses a helper depth image if present, runs it through \texttt{model.visual(...)}, and compresses the resulting sequence to the fixed latent count (default \texttt{latent\_size=4}) by averaging groups. The count-matched GT variant uses a two-pass protocol: first generate once, count latent tokens, then recompress GT latents to that count and regenerate.

\paragraph{Mirage training data and latent collation.}
Mirage's depth training utilizes a dataset of 19{,}279 examples (\texttt{train\_depth\_long.jsonl}) for both stages of training. In the JSONL schema, the \texttt{text\_input} contains the point-localization question and step-by-step reasoning prompt, while the \texttt{text\_output} contains the rationale and final answer. The \texttt{image\_output} key points to the ADE depth map.

Mechanically, Mirage does not store latent tokens directly in the JSONL. Instead, during data collation, the raw row is converted into a standard user/assistant turn. The assistant's output image slot is dynamically converted into a latent span using \texttt{<|latent\_start|>} and \texttt{<|latent\_end|>}, which is then expanded into a fixed sequence of \texttt{<|latent\_pad|>} tokens. Our default Mirage training scripts use a sequence length of \texttt{latent\_size=4}.

\subsection{Mull-Tokens}
\label{app:mull-repro}

Our Mull-Tokens reproduction uses \texttt{array/\allowbreak Qwen2.5-VL-Mull} and \texttt{array/\allowbreak Qwen2.5-VL-MullGRPO}. The prompt template appends an assistant message containing
\begin{quote}
  \texttt{"<think>" + "<|latent\_pad|>"*20 + "</think>"}.
\end{quote}
Answers are extracted from \texttt{<answer>...</answer>}.

Mull ablations replace both prefill \texttt{<|latent\_pad|>} embeddings and autoregressive latent embeddings after \texttt{<|latent\_start|>}. The \emph{same-dist random} setting samples each replacement vector independently from $\mathcal{N}(\mu_i, \sigma_i)$, where $\mu_i$ and $\sigma_i$ are computed from that vector's own statistics. GT and GT-dist modes require an external helper-image directory.

\subsection{CoVT}
\label{app:covt-repro}

Our CoVT reproduction uses four VLMEvalKit-wrapped checkpoints under \texttt{Wakals/}: \texttt{CoVT-7B-depth}, \texttt{CoVT-7B-\allowbreak seg\_depth\_dino}, \texttt{CoVT-7B-\allowbreak seg\_depth\_dino\_edge}, and \texttt{CoVT-LLaVA-13B-depth}.

At evaluation time, ablations replace the anchor-pad token embeddings associated with tokens such as \texttt{<|sam\_pad|>}, \texttt{<|dino\_pad|>}, and \texttt{<|depth\_pad|>}. Supported modes are
\[
  \{\texttt{zero}, \texttt{random}, \texttt{same}, \texttt{random\_dist}, \texttt{first\_repeat}\}.
\]
GT and count-matched modes are not implemented in the CoVT evaluation wrappers.

% ========================================== %
% Appendix G: Full Results and Variance
% ========================================== %
% \section{Full Results and Variance}
% \label{app:variance}

% The repository contains aggregate evaluation files such as \texttt{all\_evaluation\_results\_sorted.txt} and \texttt{all\_evaluation\_results.tsv}, as well as per-run outputs in \texttt{evaluation\_summary*.json}. There is also explicit five-run / sequential infrastructure under \texttt{checkpoints/seq\_once\_5runs/...} and \texttt{answers/seq\_once\_5runs/...}.

% At present, we did not find a script that automatically computes variance or error bars from these repeated runs. A table-ready summary script remains to be added:
% \begin{quote}
%   \textbf{TODO:} script for variance / error-bar aggregation across repeated runs.
% \end{quote}

% ========================================== %
% Appendix H: Compute and Resources
% ========================================== %
\section{Compute and Resources}
\label{app:compute}

The compute settings used in our experiments are summarized below:
 1 node on partition \texttt{gpu-l40s}, 8 GPUs, 40 CPUs per task, and 300\,GB RAM. 10 epochs of continuous training takes approximately 10 wall-clock hours.

\end{document}